\documentclass[conference]{IEEEtran}
\IEEEoverridecommandlockouts
\usepackage{cite}
\usepackage{amsmath,amssymb,amsfonts}
\usepackage{algorithmic}
\usepackage{graphicx}
\usepackage{textcomp}
\usepackage{xcolor}
\usepackage{hyperref}
\usepackage{graphicx}
\usepackage{booktabs}
\usepackage{multirow}
\def\BibTeX{{\rm B\kern-.05em{\sc i\kern-.025em b}\kern-.08em
    T\kern-.1667em\lower.7ex\hbox{E}\kern-.125emX}}
\begin{document}

\title{5W1H Extraction With Large Language Models}
% \author{\IEEEauthorblockN{Anonymous Authors}}
\author{
    \IEEEauthorblockN{Yang Cao, Yangsong Lan, Feiyan Zhai, Piji Li$^{*}$\thanks{*Corresponding author.}}
    \IEEEauthorblockA{College of Computer Science and Technology,\\
Nanjing University of Aeronautics and Astronautics, China}
    \IEEEauthorblockA{MIIT Key Laboratory of Pattern Analysis and Machine Intelligence, Nanjing, China}
    \IEEEauthorblockA{\{jymmzb,lys2962331781,pjli\}@nuaa.edu.cn, \{zfy455140\}@gmail.com}
}
\maketitle

\begin{abstract}
The extraction of essential news elements through the 5W1H framework (\textit{What}, \textit{When}, \textit{Where}, \textit{Why}, \textit{Who}, and \textit{How}) is critical for event extraction and text summarization.
The advent of Large language models (LLMs) such as ChatGPT presents an opportunity to address language-related tasks through simple prompts without fine-tuning models with much time. While ChatGPT has encountered challenges in processing longer news texts and analyzing specific attributes in context, especially answering questions about \textit{What}, \textit{Why}, and \textit{How}. The effectiveness of extraction tasks is notably dependent on high-quality human-annotated datasets. However, the absence of such datasets for the 5W1H extraction increases the difficulty of fine-tuning strategies based on open-source LLMs. To address these limitations, first, we annotate a high-quality 5W1H dataset based on four typical news corpora (\textit{CNN/DailyMail}, \textit{XSum}, \textit{NYT}, \textit{RA-MDS}); second, we design several strategies from zero-shot/few-shot prompting to efficient fine-tuning to conduct 5W1H aspects extraction from the original news documents.
The experimental results demonstrate that the performance of the fine-tuned models on our labelled dataset is superior to the performance of ChatGPT. 
Furthermore, we also explore the domain adaptation capability by testing the source-domain (e.g. NYT) models on the target domain corpus (e.g. CNN/DailyMail) for the task of 5W1H extraction.
\end{abstract}


\section{Introduction}
Effective extraction of crucial news aspects using the 5W1H framework (\textit{What}, \textit{When}, \textit{Where}, \textit{Why}, \textit{Who}, and \textit{How}) is important for news aggregation, event extraction, and text summarization~\cite{hamborg2020bias,hamborg2019giveme5w1h,hamborg2017matrix,li2015reader}.
These 5W1H aspects aim to capture the most fundamental details of news that people are interested in when it comes to a specific event or topic. By absorbing these aspects, one can obtain a comprehensive understanding of the core idea of the event.
Despite the fact that 5W1H extraction is a fundamental task in news analysis, there is currently no effective method to ensure both accuracy and comprehensiveness in carrying out this task. The previous approach focused on implicitly detecting events, and often employed techniques like topic modeling. These methods typically specialize in extracting task-specific properties, such as the number of casualties in an accident.
Recent studies have shown some progress in extracting 
5W1H aspects. 
Hamborg \textit{et al.}~\cite{hamborg2018extraction} proposed Giveme5W1H, which is an open-source main event retrieval system for news articles. They first standardized the news content and then used syntax and domain-specific rules to extract phrases related to 5W1H. Finally, a scoring
mechanism was implemented for candidate statements to determine the
relevant phrases in the main event.
Giveme5W1H provides a rule-based direction for extracting 5W1H elements.
\begin{figure}[h]
    \centering
    
    \includegraphics[scale=0.49]{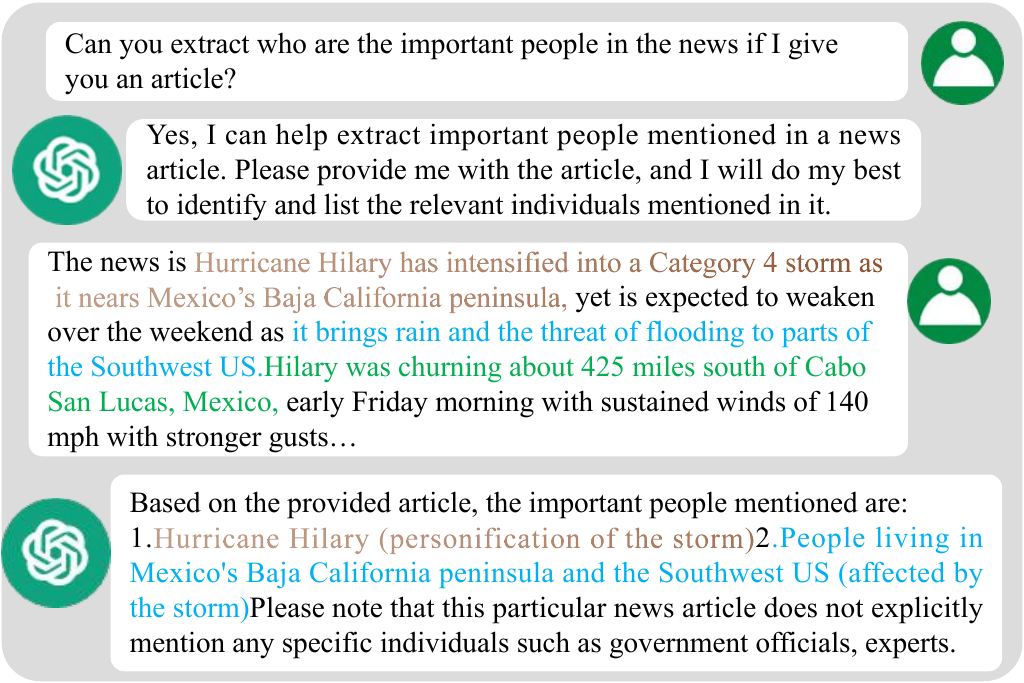}%
     % \vspace{-5pt}
     %  \vspace{-0.15cm}
	\caption{
	Different colors represent different sources of information from the original text.In the context of the article, the mention of Hillary is not meant to be a specific name, but the answer provided by ChatGPT mistakenly identifies it as one of the important people. The names of places mentioned in the green line may also refer to residential areas, but the given answers do not correspond to them.
	}
	\label{fig1}
        % \vspace{-0.05cm}
\end{figure}
For the answers to \textit{How} and \textit{Why} that involve a comprehensive understanding of the full article, however, they cannot propose a satisfied solution.

LLMs such as ChatGPT~\cite{ouyang2022training} have demonstrated remarkable capabilities especially in zero-shot/few-shot scenarios based on the prompting or in-context leaning strategies~\cite{gao2023text,chen2023improving}.
However, 5W1H extraction is a more intricate task that presents unique challenges. In contrast to tasks like summarization, 5W1H extraction necessitates a more detailed and precise description of the content.
To use ChatGPT for 5W1H extraction, it is crucial to provide a comprehensive and detailed task description and select an appropriate prompt as the questioning method. Additionally, handling complex special cases is critical for accurate extraction.
Nevertheless, in our investigations we observe that due to the limitations of context length, their abilities of 5W1H extraction from full news reports are particularly insufficient.
As shown in Figure~\ref{fig1}, we input ChatGPT with a news article and request it to extract aspect \textit{Who} from the text. 
From the results we can find that ChatGPT can extract named entities from the text, but it cannot analyze their specific types within the context. ``Hilary'' is not a commonly mentioned person name in the article, but rather the name of a hurricane. In addition, in terms of the comprehensiveness of information, ChatGPT ignores the content described in the green highlighted section of the text. Therefore, solely relying on the capabilities of LLMs themselves is far from enough.

Fine-tuning a model with task-specific human-created training data can achieve performance improvement~\cite{gao2023exploring,yang2019exploring,lou2022translation}. Therefore, the quality and quantity of samples in the annotated dataset are crucial for the task.
RA-MDS\cite{li2017reader} is a dataset consisting of manually annotated news event aspects, which contains six different categories of news and annotates the distinctive aspects for each category.
However, it did not extract 5W1H elements for all categories, resulting in the loss of certain category attributes. Additionally, the number of articles is relatively small, which may not be sufficient for effectively fine-tuning LLMs.
In addition, the length of the news articles is typically quite long, which increases the memory requirement for training. Thus efficient fine-tuning approaches are required due to the limitations of computing resources. Therefore the extraction of 5W1H elements becomes considerably more challenging.

To address the above problems,  first, we annotate four publicly accessible news datasets(\textit{CNNDM}, \textit{XSum}, \textit{NYT}, and \textit{RA-MDS}) to extract statements from the text that are used in 5W1H questions. In total, we labeled 3,500 entries within these datasets. Second, for the fine-tuning techniques, recently, some efficient methods of fine-tuning LLMs are proposed, such as Prefix-Tuning\cite{li2021prefix}, Low-Rank Adaptation (LoRA)\cite{hu2021lora}, and QLoRA\cite{dettmers2023qlora}. They significantly reduce the resources and time required for fine-tuning the model. Then we fine-tune these LLMs on the annotated dataset and assess their performance in extracting 5W1H elements. Additionally, we also conduct an evaluation of the capability of news across different domains.

\vspace{-1.5pt}
Our contributions are as follows: 
 1) We annotate a high-quality 5W1H elements extraction dataset for news in several different domains on four different datasets.  2) We fine-tune some LLMs on the annotated datasets and also conduct comparisons on zero-shot and few-shot capabilities with ChatGPT. The experimental results show that the performance of fine-tuned models is better than ChatGPT. 3) we also explore the domain adaptation capability of models by testing the source-domain (e.g. NYT) model to the target domain corpus (e.g. CNN/DailyMail) for the task of 5W1H extraction. 
\section{Related Work}

\subsection{Event extraction}
5W1H elements extraction from news articles is a specific area of event extraction. However, recent research has not used LLMs for extracting 5W1H information.

Hamborg et al.\cite{hamborg2018giveme5w} proposed the first open-source system (\textbf{Giveme5W}) for retrieving the main event from news articles. The system uses syntactic and domain-specific rules to extract and score phrase candidates for each 5W question. Answering \textit{How} questions becomes particularly complex due to the following reasons. First, the answer to the \textit{How} question usually depends on the context and may require a comprehensive understanding of the relevant information in the text. This may require the system to search for information in a wide range of contexts, rather than relying solely on the current sentence or paragraph. Second, It may be influenced by ambiguity, as an event or process may have multiple interpretations.
The system needs to be able to recognize and handle this ambiguity in different contexts. In addition, the \textit{How} question is considered less important in many use cases when compared to other questions\cite{das20125w,parton2009comparing,yaman2009classification}. Most systems focus only on the extraction of 5W questions without \textit{How} questions.
Hamborg et al.\cite{hamborg2019giveme5w1h} proposed the framework Giveme5W1H based on Giveme5W. They addressed \textit{How} questions using semantic and grammatical rules and tested their effectiveness on multiple datasets. The extraction of content related to \textit{How} and \textit{Why} questions, requiring a thorough understanding of the entire text, is notably insufficient when relying solely on semantic rules.

Recent studies aim to understand events and capture their correlations by using LLMs. Gao et al.\cite{gao2023exploring} used ChatGPT and task-specific modeling to compare performance on the event extraction task and explored the effects of different prompts on ChatGPT. However, their experimental results indicated that despite attempting numerous prompts, ChatGPT still did not perform as well as task-specific models. Srivastava et al.\cite{srivastava2023mailex}presented the first dataset, MAILEX, for performing event extraction from conversational email threads. They explored the impact of ChatGPT on the email event extraction task in a few-shot scenario. But the experimental results showed that in-context learning in ChatGPT demonstrated poorer performance compared to the fine-tuned approaches. Li et al.\cite{li2023evaluating} found that ChatGPT's performance in the standard information extraction setting is not as effective as that of models based on the BERT-based models in most cases. We find that there is currently no research on 5W1H elements in news extraction using LLMs. So we use LLMs to train a high-quality annotated dataset.
\subsection{Fine-tuing LLMs}
With the advent of LLMs, many fine-tuning methods have emerged. Instruction tuning\cite{wei2021finetuned} is a method through which pretrained autoregressive language models are finetuned to follow natural language instructions and generate responses. 
It helps reduce the need for annotated data, as the model can be fine-tuned through guidance without relying on large amounts of annotated task-specific data. Additionally, the model can be fine-tuned through different natural language instructions, providing greater flexibility to adapt to multiple tasks without the necessity of retraining the entire model.
Hu\cite{hu2021lora} proposed LoRA, which freezes the weights of the pretrained model, significantly reducing the number of trainable parameters and computational memory. Dettmers\cite{dettmers2023qlora} proposed QLoRA, which can replicate 16-bit full finetuning performance with a 4-bit base model. QLoRA is based on the LoRA framework, which preserves model performance while further reducing the model's memory. In this work, we use QLoRA to fine-tune LLMs. 

\begin{figure}[htp]  
\centering  
 \includegraphics[scale=0.38]
 {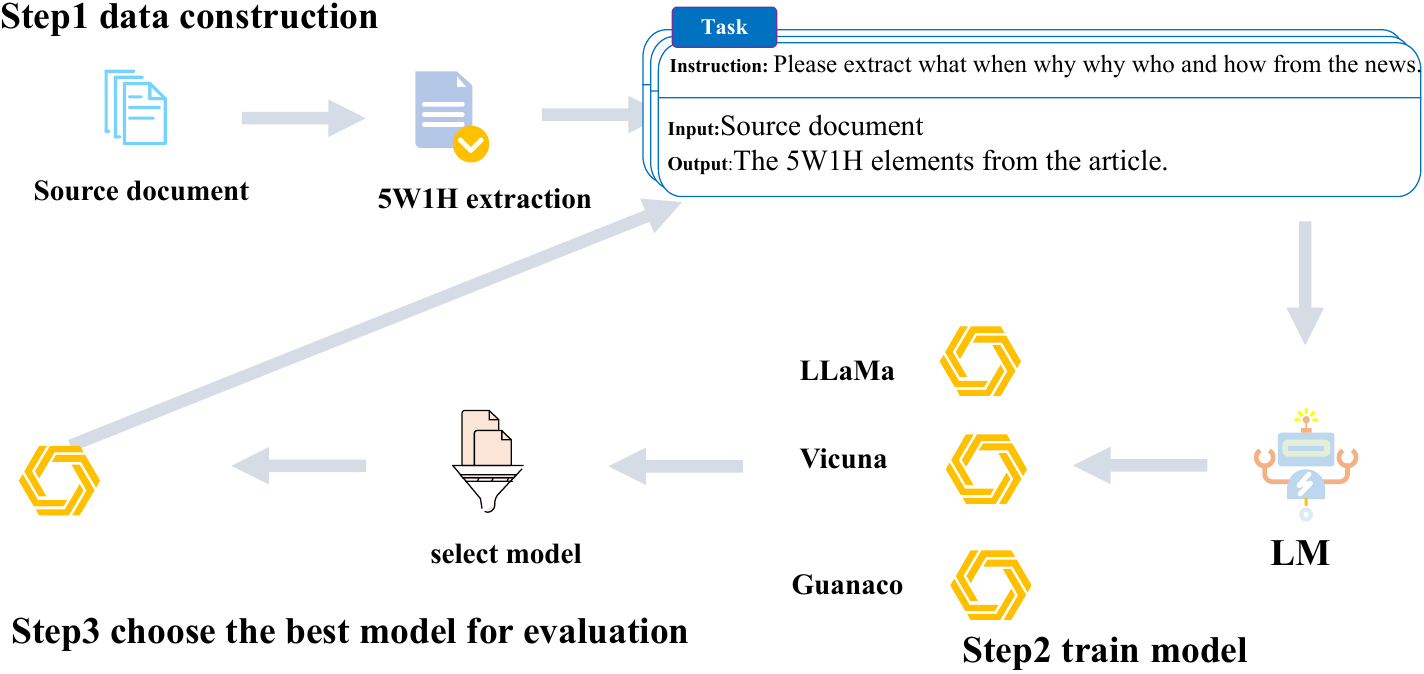}
\caption{\textbf{Fine-tuning LLMs on 5W1H elements extraction from news articles.} Construct annotated datasets, convert them into the required format for input into LLMs, fine-tune with different models, and select the best model based on evaluation results.}  
\label{figure2}
\end{figure}

\section{METHODOLOGY}
In this section, we describe the process of constructing a dataset for extracting 5W1H elements from news and explain how to use LLMs to fine-tune the constructed dataset.
The architecture is shown in Figure~\ref{figure2}, which consists of three parts. First, we extract 5W1H elements from news articles by professional annotators. Subsequently, we process the annotated dataset into Supervised Fine-Tuning (SFT) format.
Next, we choose three models to train the dataset and evaluate their effectiveness. Finally, we select the model with the best performance based on the evaluation results.

\subsection{Dataset Construction}
We annotate the 5W1H dataset based on four typical news corpora (CNN/DailyMail, XSum, NYT, RA-MDS). In total, we label 3,500 entries within these datasets.
We follow Li's \textit{et al.}\cite{li2017reader} categorization of news during dataset construction.
There are 6 predefined categories in the labeled news: (1) Accidents and Natural Disasters, (2) Attacks (Criminal/Terrorist), (3) New Technology, (4) Health and Safety, (5) Endangered Resources, and (6) Investigations and Trials (Criminal/Legal/Other).
A variety of different elements are individually defined within each category.
In contrast to the Li's \textit{et al.}\cite{li2017reader}
approach, we have added elements of 5W1H to each category.
\vspace{-0.5pt}
We ask the experts to determine which category the article belongs to based on its content. After determining the category of the article, we analyze its content and extract all the statements related to 5W1H from the original text.

Annotators are primarily engaged in AI data annotation services, with particular expertise in English business scenarios. The team is one of the suppliers that has been cooperating with Tencent for a long time, and due to the confidentiality agreement, we have not provided specific information about the team. Approximately 10 individuals are dedicated to this task.
Each article is collaboratively annotated by a team of three annotators. For the short entities mentioned in the article, such as \textit{Who}, \textit{Where}, and \textit{When}, we can directly extract information from the original text. However, for long entities like \textit{What}, \textit{Why}, and \textit{How}, a comprehensive understanding of the article's content is required to better extract relevant information. We have made the following attempts to extract \textit{Why} from the original text. We ask the annotators to identify all events in the article, each of which may involve multiple reasons, and identify each corresponding reason from the text one by one. Finally, the third annotator will double-check for any omissions. The same applies to \textit{What} and \textit{How}. Before the formal annotation process, we select 50 news articles from each dataset for preliminary annotation by our annotators. Detailed instructions are provided to address any issues encountered during this preliminary phase. The average token count for articles in each dataset is presented in Table \ref{tab:addlabel}. In order to facilitate labeling by experts we specifically choose articles with an approximate length of 550 words.

\begin{figure}[htp]  
\centering  
 \includegraphics[scale=0.43]{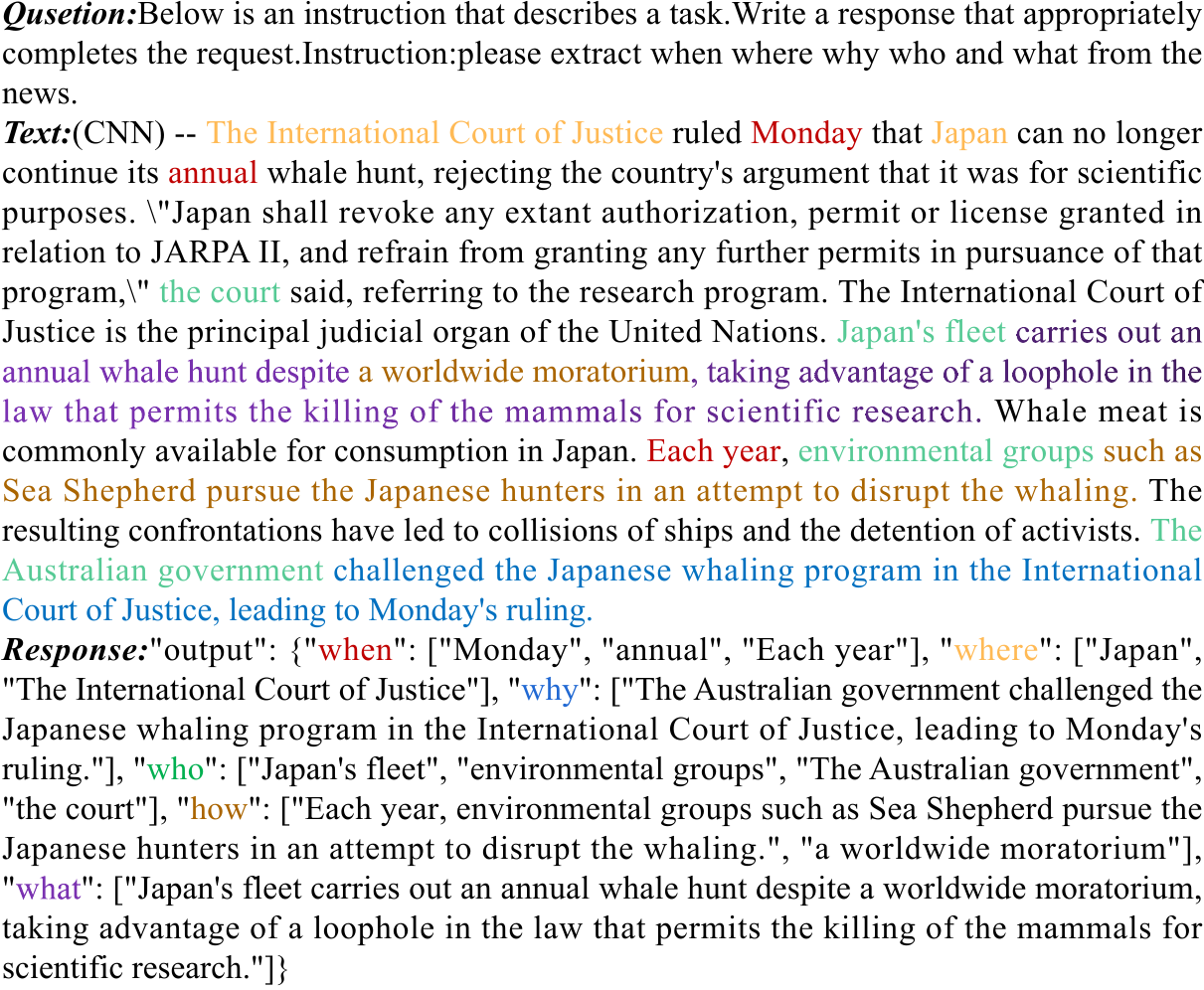}
\caption{Correspondence between 5W1H aspects and the original text, with different colors representing different answers to questions.}  
\label{figure3}
\end{figure}

% Table generated by Excel2LaTeX from sheet 'Sheet1'
\begin{table}[htbp]
  \centering
   \caption{The average words of articles in each dataset before formal annotation.}
    \begin{tabular}{lc}
    \toprule
    Dataset & \multicolumn{1}{l}{AvgDocWords} \\
    \midrule
    CNN/DailyMail & 579 \\
    XSum  & 523 \\
    NYT   & 552 \\
    RA-MADS & 568 \\
    \bottomrule
    \end{tabular}%
      % \caption{The average words of articles in each dataset before formal annotation.}
  \label{tab:addlabel}%
\end{table}%
For the accuracy of the answers, all marked elements are directly extracted from the original text without any semantic generalization or paraphrasing.
In the labeling process, for each statement (if it belongs to the 5W1H elements), there must be a uniquely identified element corresponding to it. For instance, if the statement is categorized as \textit{What}, it cannot be classified as \textit{How} or any other case. This approach helps prevent ambiguity in labeling statements.
To ensure that key information is not omitted, we annotate all relevant 5W1H elements in each article during the labeling process. For instance, in a news article, there might be multiple instances of time and location mentioned, and an event could have various reasons. We conduct comprehensive extraction for each of these elements to cover all potential key information present in the article.
At the same time, to prevent redundancy in labeling each sentence, we choose to extract a fragment of the statement (some words). For example, a sentence with a brief mention of the time and place (only about nouns) could be extracted without including the entire sentence. 
During the labeling process, there are certain types of news where identifying the corresponding element proves challenging. For the identification of some elements, we adopt a broader definition based on the contextual meaning of the original text. In some technology-related articles, there may not be explicit mentions of location-related information. Instead, scenarios could involve discussing the transmission of information through the medium of the internet. Therefore, in such cases, we define the internet as a more abstract location.
We try to label each article with 5W1H, but there might be instances where we are unable to extract all the elements of 5W1H from certain news articles.
Finally, we save the labeled datasets in JSON format. The annotated content of the 5W1H elements from the article is shown in Figure~\ref{figure3}.

\subsection{Fine-tuning for 5W1H Extraction}
After obtaining the annotated dataset, we apply instruction-tuning to LLMs to compare the differences between our annotated dataset and the results extracted using ChatGPT. Our template is shown in Figure ~\ref{figure3}. The LLMs can directly comprehend the 5W1H methodology and extract relevant content from the original text. Therefore, we don’t employ the Chain-of-Thought (CoT) approach\cite{wei2022chain}, which involves guiding the LLMs through a conversational method to understand and retrieve content for the extraction task. Instead, we instruct the LLMs to directly extract content related to 5W1H from the original text. 
In this approach, the model is presented with articles and instructions as input and the 5W1H elements are used as the output of the model. We input the original text and instructions into the LLMs, and the model will generate the extracted information of the 5W1H elements from the article. More specifically, the LLMs are presented with an article \emph{X}$_{\mbox{\scriptsize passage}}$ as input. For each 5W1H element \emph{E}$_{\mbox{\scriptsize elements}}$ that appears in the output, we convert them into a natural language query, such as "Please extract \textit{What}, \textit{When}, \textit{Where}, \textit{Why}, \textit{Who}, and \textit{How} from the news." Subsequently, we fine-tune the LLMs to generate a structured output \emph{y}$_{\mbox{\scriptsize i}}$  in the format of a JSON list containing all entities \emph{E}$_{\mbox{\scriptsize elements}}$. \emph{y}$_{\mbox{\scriptsize i}}$ represents the collection of extracted 5W1H elements from each article.

We choose \textbf{LLaMA}\footnote{The checkpoint is “huggyllama/llama”.}\cite{touvron2023llama}, \textbf{Vicuna}\footnote{The checkpoint is “eachadea/vicuna”.}\cite{chiang2023vicuna},
and \textbf{Guanaco}\footnote{The checkpoint is “TheBloke/guanaco”.}\cite{dettmers2023qlora} as training models and employ QLoRA fine-tuning approach, following the training schedule of Dettmers \textit{et al.} To assess the impact of different model sizes, we evaluate the performance of 7B and 13B models separately. Before fine-tuning with LLMs, we treat the task of extracting 5W1H elements from the news as a standard Named Entity Recognition (NER)  task, given its close association with specific terms and NER-related challenges.
News, being a form of lengthy textual content, poses a challenge when we choose the \textbf{XLNet}\cite{yang2019xlnet} pre-trained model for processing. However, the experimental outcomes are less than satisfactory, especially in the extraction of long entities, where the results are nearly non-existent. This indicates the inadequacy of relying solely on pre-trained models for effective 5W1H element extraction. Meanwhile, this also indicates that the NER task is not suitable for this specific task. Hence, the appropriate evaluation metrics for assessing the effectiveness of 5W1H element extraction becomes crucial.

During the inference stage, to enhance the model's inference speed, we utilize the \textbf{CTranslate2}\cite{klein2020opennmt}framework to convert the combined LORA weights from the fine-tuned model and the base model into the CTranslate2 format. We reduce the elapsed time by employing int8 quantization, which results in a slight effect on the performance of the model. However, this will not have a significant impact on the inference performance of the model. We select the best-performing model from the fine-tuned models based on evaluation metrics to apply to other datasets. Finally, we compare fine-tuned models with \textbf{ChatGPT} and \textbf{GPT-4} and save the results in JSON format. 

\section{EXPERIMENTS}
In this section, we describe the experiment details including datasets, implementation details, and the results.
\subsection{Datasets}
\textbf{CNN/DailyMail}\cite{nallapati2016abstractive} serves as a single text summarization corpus, which is commonly used for extractive summarization tasks. It includes summaries with multiple sentences and is collected from CNN and DailyMail.

\textbf{XSum}\cite{narayan2018don}dataset is a highly abstractive dataset and consists of single-sentence summaries. It is particularly suitable for evaluating models in tasks requiring concise abstractive summarization.

\textbf{NYT}\cite{sandhaus2008new}dataset comprises millions of articles from The New York Times, spanning 1987-2007. It includes over 650,000 staff-written summaries and 1.5 million annotated articles, suitable for tasks like automatic summarization and text classification.

\textbf{RA-MDS}\cite{li2017reader}is a multi-document news summary dataset that focuses on extracting news comments. The articles highlight readers' points of interest in news comments, providing valuable perspectives for summary systems. This dataset encompasses 450 news articles from six different domains.

The first three datasets each contain 1000 instances. Each dataset is divided according to 8:1:1 for training, validation, and testing. Since the RA-MDS dataset contains 450 news, we decide not to divide it separately. Instead, we merge this part of the dataset into the first three datasets for training.
\begin{table*}[t]
   \caption{Each element in the experimental results has valid responses of over 80; responses with fewer than 80 are not displayed in the results. R-2 represents Rouge-2, R-L represents Rouge-L and B-4 represents BLEU-4}
  \centering
  \small

  \resizebox{2\columnwidth}{!}{
    \begin{tabular}{cc|ccc|ccc|ccc|ccc|ccc|ccc}
    \toprule
   \multicolumn{19}{c}{\textit{\textbf{CNN/DailyMail}}} \\
    \midrule
    \multicolumn{2}{c|}{\multirow{2}[2]{*}{\textbf{Model}}} & \multicolumn{3}{c|}{\textbf{What}} & \multicolumn{3}{c|}{\textbf{Who}} & \multicolumn{3}{c|}{\textbf{Why}} & \multicolumn{3}{c|}{\textbf{When}} & \multicolumn{3}{c|}{\textbf{Where}} & \multicolumn{3}{c}{\textbf{How}} \\
    \multicolumn{2}{c|}{} & R-2   & R-L   & B-4   & R-2   & R-L   & B-4   & R-2   & R-L   & B-4   & R-2   & R-L   & B-4   & R-2   & R-L   & B-4   & R-1   & R-2   & B-4 \\
    \hline
    \multicolumn{2}{c|}{\textit{Fine-tuing}} &       &       &       &       &       &       &       &       &       &       &       &       &       &       &       &       &       &  \\
    \multicolumn{2}{c|}{LLaMa-7B} & \textbf{39.20} & \textbf{47.67} & \textbf{55.75} & 3.59  & 19.42 & 58.61 & 17.05 & 25.49 & 34.14 & 17.12 & 31.16 & \textbf{62.97} & 9.49  & 19.76 & 54.07 & 14.42 & 24.57 & 28.09 \\
    \multicolumn{2}{c|}{Vicuna-7B} & —     & —     & —     & 2.88  & 18.42 & \textbf{64.81} & 14.83 & 22.07 & 29.62 & 14.31 & 28.92 & 62.63 & 6.83  & 18.17 & \textbf{57.90} & 14.60 & 24.60 & 35.62 \\
    \multicolumn{2}{c|}{Guanaco-7B} & 26.51 & 36.19 & 28.78 & 3.39  & 19.90 & 51.70 & \textbf{17.60} & \textbf{26.16} & 24.26 & 11.98 & 25.77 & 56.17 & 8.30  & 19.53 & 51.51 & 16.22 & 25.73 & 19.36 \\
    \multicolumn{2}{c|}{LLaMa-13B} & 32.75 & 42.30 & 32.37 & 5.16  & 21.81 & 52.64 & 13.61 & 23.59 & 28.42 & 15.56 & 32.49 & 49.00 & 11.27 & 23.53 & 45.64 & 21.96 & 31.93 & 31.34 \\
    \multicolumn{2}{c|}{Vicuna-13B} & 31.25 & 40.61 & 29.53 & 4.71  & 20.66 & 50.45 & 13.91 & 23.41 & 30.08 & 12.35 & 28.07 & 49.87 & 6.43  & 17.27 & 41.77 & \textbf{22.41} & \textbf{32.35} & 31.98 \\
    \multicolumn{2}{c|}{Guanaco-13B} & 28.85 & 38.38 & 29.20 & 4.06  & 20.81 & 50.69 & 14.54 & 24.69 & 24.04 & \textbf{17.98} & 36.18 & 46.50 & 14.61 & 27.34 & 43.23 & 21.14 & 31.48 & 27.29 \\
    \hline
    \multicolumn{2}{c|}{\textit{zero-shot}} &       &       &       &       &       &       &       &       &       &       &       &       &       &       &       &       &       &  \\
    \multicolumn{2}{c|}{GPT-3.5-turbo} & 13.25 & 28.15 & 46.19 & 3.75  & 11.75 & 26.84 & 10.80 & 21.42 & 33.71 & 3.76  & 11.91 & 21.44 & 10.04 & 19.18 & 30.23 & 7.86  & 21.29 & 35.67 \\
    \multicolumn{2}{c|}{GPT-4} & 16.44 & 29.53 & 52.04 & 3.36  & 9.87  & 22.91 & 10.50 & 22.07 & 29.76 & 2.88  & 7.26  & 15.79 & 3.98  & 9.99  & 22.70 & 7.88  & 20.45 & 35.31 \\
    \hline
    \multicolumn{2}{c|}{\textit{few-shot}} &       &       &       &       &       &       &       &       &       &       &       &       &       &       &       &       &       &  \\
    \multicolumn{2}{c|}{GPT-3.5-turbo} & 12.29 & 22.80 & 48.72 & 3.08  & 10.12 & 27.26 & 8.75  & 17.68 & 30.79 & 6.69  & 15.27 & 23.53 & 7.84  & 16.69 & 28.59 & 5.81  & 15.66 & 30.02 \\
    \multicolumn{2}{c|}{GPT-4} & 35.61 & 45.33 & 52.82 & \textbf{5.57} & \textbf{19.60} & \textbf{46.78} & 16.51 & 25.19 & \textbf{34.88} & 17.03 & 31.17 & 45.22 & \textbf{15.98} & \textbf{27.85} & 46.83 & 12.15 & 21.00 & 35.86 \\
     \midrule
      \multicolumn{19}{c}{\textit{\textbf{XSum}}} \\
      \midrule
    \multicolumn{2}{c|}{\multirow{2}[2]{*}{\textbf{Model}}} & \multicolumn{3}{c|}{\textbf{What}} & \multicolumn{3}{c|}{\textbf{Who}} & \multicolumn{3}{c|}{\textbf{Why}} & \multicolumn{3}{c|}{\textbf{When}} & \multicolumn{3}{c|}{\textbf{Where}} & \multicolumn{3}{c}{\textbf{How}} \\
    \multicolumn{2}{c|}{} & R-2   & R-L   & B-4   & R-2   & R-L   & B-4   & R-2   & R-L   & B-4   & R-2   & R-L   & B-4   & R-2   & R-L   & B-4   & R-1   & R-2   & B-4 \\
    \hline
    \multicolumn{2}{c|}{\textit{Fine-tuing}} &       &       &       &       &       &       &       &       &       &       &       &       &       &       &       &       &       &  \\
    \multicolumn{2}{c|}{LLaMa-7B} & 41.74 & 51.19 & 53.96 & 8.00  & 19.49 & 44.70 & 16.83 & 25.59 & 36.77 & 11.39 & 27.07 & 43.81 & 7.59  & 17.77 & 42.74 & \textbf{22.63} & \textbf{32.34} & 41.67 \\
    \multicolumn{2}{c|}{Vicuna-7B} & —     & —     & —     & 9.56  & 21.53 & 45.37 & 18.18 & 27.97 & 35.92 & 11.28 & 27.45 & 46.97 & 7.48  & 17.65 & 43.30 & 19.55 & 29.76 & 36.41 \\
    \multicolumn{2}{c|}{Guanaco-7B} & 42.07 & 51.41 & 48.41 & 10.33  & 21.18 & 42.13 & 16.73 & 25.36 & 32.00 & 10.53 & 26.75 & 43.14 & 7.34  & 17.63 & 40.97 & \textbf{20.22} & 30.05 & 35.01 \\
    \multicolumn{2}{c|}{LLaMa-13B} & 44.32 & 53.36 & 52.04 & 7.49  & 18.40 & 47.49 & 17.24 & 26.61 & 38.19 & 10.58 & 27.67 & 46.81 & 8.82  & 19.89 & 43.29 & 15.52 & 25.19 & 36.50 \\
    \multicolumn{2}{c|}{Vicuna-13B} & 40.92 & 50.62 & 54.83 & 8.18  & 21.34 & 45.18 & 20.22 & \textbf{29.45} & \textbf{40.57} & 11.35 & 27.44 & 45.27 & 7.60 & 17.25 & 41.81 & 19.38 & 28.65 & 38.51 \\
  
    \multicolumn{2}{c|}{Guanaco-13B} & 43.20 & 52.61 & 49.81 & 9.18  & 21.66 & 45.49 & 14.29 & 23.26 & 32.79 & 11.00 & 28.12 & 45.62 & 8.32 & 20.14 & 43.08 & 17.69 & 28.21 & 38.20 \\
    \hline
    \multicolumn{2}{c|}{\textit{zero-shot}} &       &       &       &       &       &       &       &       &       &       &       &       &       &       &       &       &       &  \\
    \multicolumn{2}{c|}{GPT-3.5-turbo} & 10.60 & 22.97 & 43.06 & 7.43  & 16.98 & 30.31 & 8.17 & 18.42 & 35.75 & 6.18  & 13.70 & 20.62 & 7.56 & 16.72 & 29.03 & 9.54  & 19.72 & 37.89 \\
    \multicolumn{2}{c|}{GPT-4} & 13.43 & 26.77 & 44.04 & 8.81  & 20.81  & 35.88 & 10.04 & 20.62 & 34.29 & 7.30  & 17.70  & 25.97 & 8.65  & 19.89  & 38.57 & 10.50  & 21.68 & 38.43 \\
    \hline
    \multicolumn{2}{c|}{\textit{few-shot}} &       &       &       &       &       &       &       &       &       &       &       &       &       &       &       &       &       &  \\
    \multicolumn{2}{c|}{GPT-3.5-turbo} & —  & —  & —  & 3.95  & 9.98 & 16.61 & 8.39  & 17.17 & 22.74 & 1.79  & 4.96 & 3.58 & 1.92  & 6.08 & 6.31 & 6.99  & 16.45 & 36.19 \\
    \multicolumn{2}{c|}{GPT-4} & \textbf{44.99} & \textbf{54.24} & \textbf{61.58} & \textbf{13.09} & \textbf{23.31} & \textbf{52.01} & 19.02 & 27.78 & 39.22 & \textbf{18.87} & \textbf{34.08} & \textbf{57.62} & \textbf{10.05} & \textbf{21.83} & \textbf{50.44} & 17.74 & 25.32 & \textbf{47.24} \\
    \bottomrule
    \end{tabular}%
    }
    % \caption{Each element in the experimental results has valid responses of over 80; responses with fewer than 80 are not displayed in the results. R-2 represents Rouge-2, R-L represents Rouge-L and B-4 represents BLEU-4}
  \label{tab:addlabel2}%
\end{table*}%
% Table generated by Excel2LaTeX from sheet 'Sheet1'
% \begin{table}[htbp]
%   \centering
%     \begin{tabular}{lcc}
%     \toprule
%     Dataset & \multicolumn{1}{l}{MaxInputToken} & \multicolumn{1}{l}{MaxOutputToken} \\
%     \midrule
%     CNN/DailyMail & 600   & 600 \\
%     XSum  & 600   & 600 \\
%     NYT   & 600   & 600 \\
%     \bottomrule
%     \end{tabular}%
%   \label{tab:addlabel1}%
%   \caption{The maximum input length and maximum output length of different datasets are input into the model. InputToken represents instructions and text and  OutputToken represents extracted 5W1H elements}
% \end{table}%

\subsection{Implementation details}
To adapt to the task of extracting 5W1H elements, we adjust the training strategy for the Dettmers \textit{et al.} aspect. We truncate the length of each news article to 750 tokens. The maximum input length and output length of the article are determined based on different datasets. We set source\_max\_len and target\_max\_len based on different datasets and set max\_new\_tokens to 1024. We save the checkpoint every 500 steps and conduct model training for 1000 steps. In the decoding phase, we employ the Top-p sampling strategy. In Top-p sampling, the model selects the next word based on the probability distribution but only stops when the cumulative probability reaches a certain threshold. Top-p sampling can be used to balance the diversity and controllability of generated text. 
We set the sampling\_topp to 0.95 and sampling\_temperature to 0.7. To better ensure that the text output retains as much relevant information about the 5W1H elements as possible, we set max\_length to 2000.

We don't follow the evaluation rules and procedure as described by Hamborg \textit{et al.}\cite{hamborg2018giveme5w}They asked three assessors to judge the relevance of each answer on a 3-point scale (non-relevant, partially relevant, and relevant) and used mean average generalized precision (MAgP) as a measure of 5W1H question answering.
We think that people may have varying opinions on the answers to the questions, and it is possible that this scoring mechanism could influence the final results.
Additionally, the extraction for the 5W1H elements task is different from the event extraction task. It involves extracting not only short entities like time, location, and people but also corresponding sentences for long entities from the original text. However, event extraction tasks do not evaluate sentences. We use the text-generation evaluation metrics \textbf{ROUGE}\cite{lin2004rouge} and \textbf{BLEU}\cite{papineni2002bleu}. The extracted 5W1H information generated by the LLMs is compared with the labeled datasets to determine the level of similarity. 
In the following experiments, we adopt ROUGE-1, ROUGE-2, and BLEU-4 for evaluation.

All experiments are conducted on NVIDIA RTX3090 GPUs (24G memory). For the 7B model size, we train using a single NVIDIA RTX 3090 GPU for approximately 13 hours. On the other hand, for the 13B model size, we use two NVIDIA RTX3090 GPUs and train for about 24 hours. Both of these training sessions are conducted on a single dataset.
\begin{table*}[t]
  \centering
   \caption{The results on XSum and NYT.R-2 represents Rouge-2,R-L represents Rouge-L and B-4 represents BLEU-4}
%\setlength\tabcolsep{1.1pt}
%  \renewcommand{\arraystretch}{1.1}
  % \caption{The results on XSum and NYT}
    \resizebox{2\columnwidth}{!}{
    \begin{tabular}{cc|ccc|ccc|ccc|ccc|ccc|ccc}
    \toprule
    \multicolumn{2}{c|}{\multirow{2}[2]{*}{\textbf{Model}}} & \multicolumn{3}{c|}{\textbf{What}} & \multicolumn{3}{c|}{\textbf{Who}} & \multicolumn{3}{c|}{\textbf{Why}} & \multicolumn{3}{c|}{\textbf{When}} & \multicolumn{3}{c|}{\textbf{Where}} & \multicolumn{3}{c}{\textbf{How}} \\
    \multicolumn{2}{c|}{} & R-2   & R-L   & B-4   & R-2   & R-L   & B-4   & R-2   & R-L   & B-4   & R-2   & R-L   & B-4   & R-2   & R-L   & B-4   & R-1   & R-2   & B-4 \\
    \hline
    \multicolumn{2}{c|}{XSum(Vicuna 13B)} & \textbf{40.08} & \textbf{50.22} & \textbf{55.21} & \textbf{10.64} & \textbf{26.55} & 47.86 & 16.21 & 25.42 & 35.78 & \textbf{19.87} & \textbf{39.29} & 51.54 & \textbf{10.86} & \textbf{25.15} & 47.79 & \textbf{23.00} & \textbf{32.66} & \textbf{40.00} \\
    \multicolumn{2}{c|}{XSum(CNN fine-tune)} & 35.34 & 45.01 & 37.87 & 9.26  & 24.99 & \textbf{52.84} & \textbf{21.13} & \textbf{29.74} & \textbf{35.89} & 13.66 & 30.77 & \textbf{54.71} & 9.73  & 23.71 & \textbf{52.23} & 18.99 & 28.45 & 31.20 \\
    \multicolumn{2}{c|}{NYT(Vicuna 13B)} & \textbf{41.14} & \textbf{49.05} & \textbf{46.86} & 8.65  & 24.09 & \textbf{45.25} & \textbf{20.18} & \textbf{28.97} & \textbf{34.37} & \textbf{8.15} & \textbf{19.80} & \textbf{48.74} & 8.20  & 19.15 & \textbf{45.59} & \textbf{20.05} & \textbf{29.13} & \textbf{31.98} \\
    \multicolumn{2}{c|}{NYT(CNN fine-tune)} & 30.02 & 38.24 & 27.72 & \textbf{10.43} & \textbf{24.48} & 36.39 & 19.11 & 27.71 & 31.14 & 6.01  & 17.67 & 44.30 & \textbf{8.94} & \textbf{20.23} & 44.74 & 18.36 & 27.12 & 27.17 \\
    \multicolumn{2}{c|}{XSum(Guanaco 13B)} & \textbf{44.70} & \textbf{54.19} & \textbf{54.04} & 10.47 & 25.07 & 46.14 & 14.48 & 23.89 & 32.95 & \textbf{20.08} & \textbf{38.75} & 50.89 & 12.64 & 28.33 & 47.13 & 17.20 & 27.21 & 36.36 \\
    \multicolumn{2}{c|}{XSum(CNN fine-tune)} & 36.15 & 45.32 & 37.87 & \textbf{12.45} & \textbf{27.61} & \textbf{52.84} & \textbf{19.07} & \textbf{28.42} & \textbf{35.89} & 18.00 & 36.36 & \textbf{54.71} & \textbf{13.04} & \textbf{29.50} & \textbf{52.23} & \textbf{17.40} & \textbf{27.84} & 31.20 \\
    \multicolumn{2}{c|}{NYT(Guanaco 13B)} & \textbf{42.57} & \textbf{50.64} & \textbf{42.29} & 9.16  & \textbf{26.24} & \textbf{49.82} & \textbf{19.82} & \textbf{28.51} & \textbf{38.32} & \textbf{7.94} & \textbf{24.10} & \textbf{47.89} & \textbf{10.91} & 22.12 & 47.89 & 13.18 & 23.40 & \textbf{35.49} \\
    \multicolumn{2}{c|}{NYT(CNN fine-tune)} & 29.25 & 37.74 & 20.28 & \textbf{10.00} & 22.69 & 33.72 & 18.38 & 26.54 & 30.62 & 7.41  & 22.80 & 37.43 & 10.39 & \textbf{22.21} & 35.29 & \textbf{16.42} & \textbf{24.97} & 25.93 \\
    \bottomrule
    \end{tabular}%
    }
    % \caption{The results on XSum and NYT.R-2 represents Rouge-2,R-L represents Rouge-L and B-4 represents BLEU-4}
  \label{tab:addlabel3}%
\end{table*}%
% \subsection{Models}
% \textbf{LLaMa} model is an open and efficient large-scale foundational language model released by Meta AI. The entire training dataset comprises approximately 1.4T tokens after tokenization. The training datasets used are all publicly available. The model's parameter range spans from 7B to 65B. Especially, LLaMA-13B outperforms GPT-3 (175B) on most benchmarks.

% \textbf{Vicuna} model is an open-source chatbot model trained on a dataset generated from 70K user dialogues collected from ShareGPT.com. It is fine-tuned on the LLaMa model using supervised instructions. The model's performance is evaluated using GPT-4 on outputs for 80 different questions to assess its effectiveness.

% \textbf{Guanaco} model is an instruction-tuning language model. It uses QLoRA to fine-tune the model on the OASST1 dataset and achieves good results.

\subsection{Experimental Results}
We present the results of the experiments with the labeled dataset
after fine-tuning with different LLMs. Furthermore, we assess the efficacy of 5W1H extraction from different models by analyzing the count of valid entries. Finally, we verify the compatibility of the models across diverse datasets.
\paragraph{\bf Finetuning on LLMs}
Table \ref{tab:addlabel2} shows the effect of fine-tuning different size models on extracting 5W1H elements from the CNN/DailyMail and XSum datasets. 
All experimental results are based on valid responses. 
While the 7B model demonstrates promising results for certain questions, it falls short in generating a substantial volume of valid responses, particularly when compared with the 13B model.
Due to the limitations in the model's performance, the 7B model fails to extract all the relevant 5W1H content during the inference process, resulting in a much lower count of valid responses compared to the 13B model.
Our findings suggest that responses to questions related to \textit{What}, \textit{Why}, and \textit{How} yield significantly higher results compared to those for \textit{Who}, \textit{When}, and \textit{Where}.
Understanding \textit{What}, \textit{Why}, and \textit{How} requires a thorough comprehension of the original text, leading to more detailed and intricate statements.
In other cases, it only requires extracting nouns from the article, such as time, location, and individuals, which leads to lower results.
The results further illustrate that augmenting the number of model parameters contributes to enhanced performance, suggesting a potential improvement in the model's ability to comprehend and extract 5W1H elements extraction from articles.
\paragraph{\bf Zero-shot and Few-shot on ChatGPT and GPT-4} 
\begin{figure}[!t]
    \includegraphics[scale=0.32]{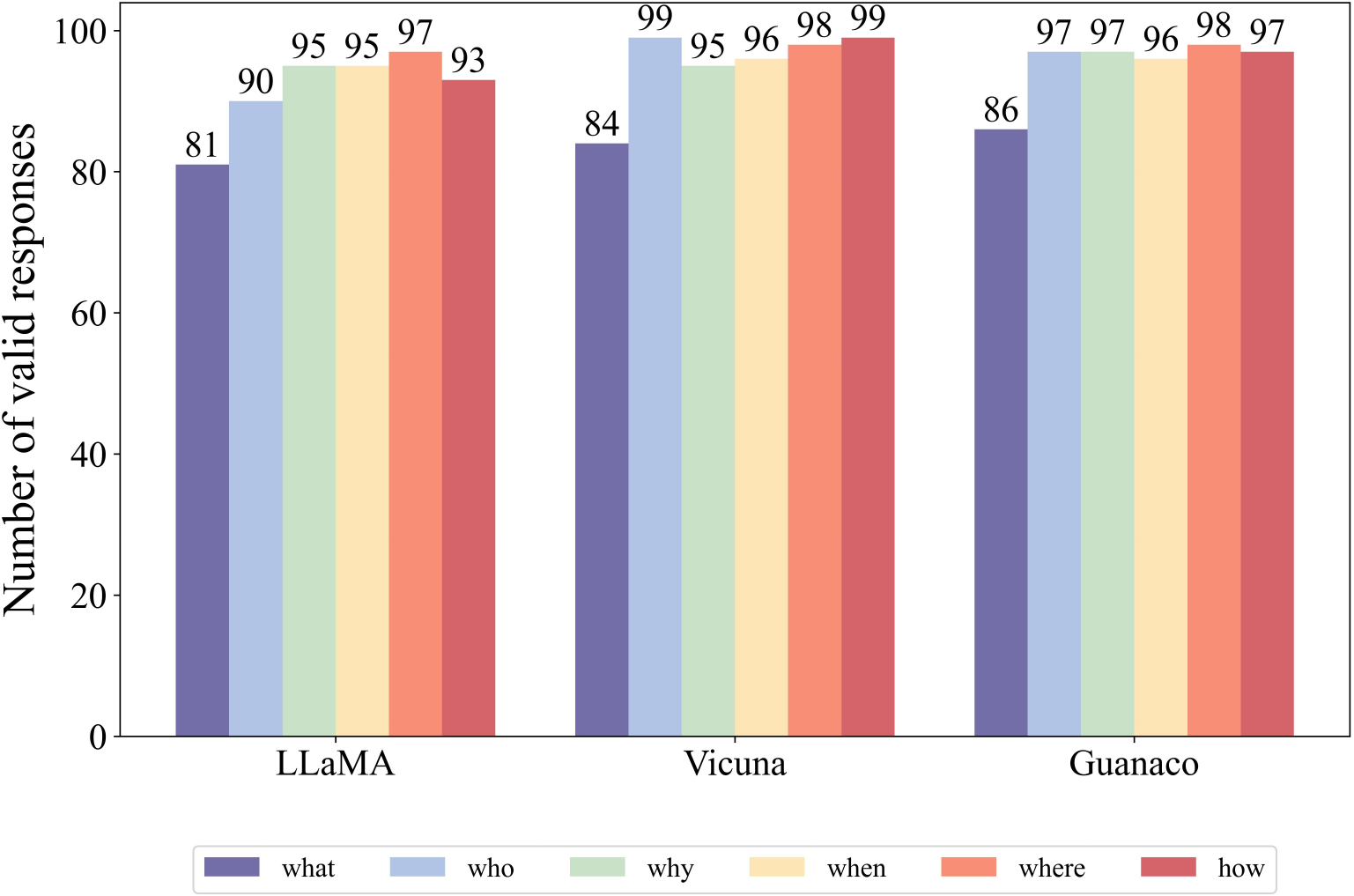}%
	\caption{
	Effective numbers of different models answering 5W1H questions on CNN/DailyMail dataset. All models are based on 13B size. Different colors represent different questions.
	}
	\label{fig3}
  \vspace{-0.4cm}
\end{figure}
We attempt to use both GPT-3.5 and GPT-4 for extracting 5W1H elements under two conditions: one without providing examples and the other with five given examples. 
The prompts are consistent with the fine-tuned LLMs. 
The experimental results indicate that without providing annotated samples, the extraction performance of both ChatGPT and GPT-4 is not satisfactory. After providing five samples, the extraction performance of ChatGPT doesn't improve. In contrast, the results of GPT-4 show a significant enhancement, especially in answering \textit{Who}, \textit{Why}, and \textit{Where} questions, where the ROUGE and BLEU were better than those after fine-tuning the LLMs.

Based on the experimental results, ChatGPT does not perform well in the task of extracting 5W1H elements. This may be due to the following reasons. News is usually long text and models may face issues of missing information. To accurately extract 5W1H elements, the model needs to have a deep understanding of the context of the text. Extracting 5W1H may involve longer entities, such as detailed information describing the process of events, which can pose challenges for the model. In contrast, GPT-4 shows significant improvement in performance after being provided with five examples.
Therefore, we believe that enhancing the model's performance in 5W1H elements extraction can be achieved by increasing the number of parameters and improving the quality of labeled datasets.
Figure \ref{fig3} illustrates the average number of 5W1H answers efficiently extracted by different models across the three datasets.
Vicuna and Guanaco have more effective answers in extracting 5W1H elements than LLaMa, and their performance is similar. Therefore we choose these two models to extract other datasets to verify their applicability in the news extraction domain.
\begin{table*}[t]
\centering
\caption{Examples of 5W1H element extraction in different situations. The 5W1H elements represented by different colored statements in the original document are annotated by professionals}
\footnotesize
  \renewcommand\arraystretch{1}
  \setlength{\tabcolsep}{2.5mm}{

\begin{tabular}{p{1.98\columnwidth}}
\toprule
\toprule
\textbf{Source Document (\textit{CNN/DailyMail})}\\
\midrule
There is no doubting the significance of \textcolor{red}{Sunday 's Tyne-Wear derby} now .\textcolor{red}{Debate has raged in Newcastle as to which their fans would rather see - a place \textcolor{blue}{in the Capital One Cup semi-final} or a first win over Sunderland in six attempts} . The majority sided with the former and a shot at a first domestic trophy \textcolor{yellow}{since 1955} .That dream , however , is dead . Alan Pardew and his players had talked of a ` massive ' week in their season . Well , it just got bigger .\textcolor{orange}{Massadio Haidara} consoles keeper \textcolor{orange}{Jak Alnwick} after a night to forget for Newcastle \textcolor{blue}{at Tottenham} .Newcastle 's defence reels as Spurs celebrate the final goal of their 4-0 defeat of the Toon \textcolor{yellow}{on Wednesday} .For Alan Pardew 's side , Sunday 's Tyne-Wear derby takes on a new dimension now they 're out of the cup .The manager had , in fairness , named just about the strongest XI available to him but it looked awfully weak against a Spurs side who were superior in every department .\textcolor{green}{The against-the-odds spirit that had inspired Newcastle 's 2-0 victory at Manchester City in the last round was missing} . It was , in the end , a rather timid surrender .Apart from Moussa Sissoko , the French battering ram who tried his best to barge his team forward , the visitors offered little .Sissoko 's compatriots Yoan Gouffran , Remy Cabella and Emmanuel Riviere -- a combined 20million worth of ` talent ' -- were passengers on a journey destined to end in defeat from the moment Jak Alnwick spilled \textcolor{orange}{Christian Eriksen} 's corner and \textcolor{orange}{Nabil Bentaleb} looped home the opening goal .A Newcastle fan , one of 4,200 supporters to travel down to London , gives his appraisal of the match .These shirtless lads do their best to lift their team having made the journey for the Capital One Cup loss .This has been a Newcastle team characterised by resistance and resilience , neither of which was in evidence .It was their second heavy loss \textcolor{blue}{in north London} \textcolor{yellow}{in five days} and such drubbings will do little for the team 's confidence come \textcolor{yellow}{Sunday} .\textcolor{orange}{Pardew} is the only manager in Newcastle 's history to lose three in a row against their North-East rivals and , with the mitigation of a cup run gone , he can ill afford another defeat \textcolor{yellow}{this weekend} .It was after Spurs ' fourth goal -- Roberto Soldado capitalising on another Alnwick spill -- that \textcolor{red}{Pardew turned his attention to the derby .Sunderland boss Gus Poyet issues instructions to his players on their way to a 1-1 draw with West Ham} .Jordi Gomez celebrates putting Sunderland 1-0 up from the penalty spot at home against West Ham .The thought of going into that game without Sissoko is a frightening one and , when he took what proved to be an innocuous knock to the knee , \textcolor{purple}{Pardew withdrew his star man almost immediately} .\textcolor{purple}{His effort was recognised by the 4,200 travelling fans and they sang until the last kick of an otherwise torrid evening} .Pardew acknowledged their support but will be in doubt about how fast such goodwill would evaporate should they fail to show against Gus Poyet 's side .The debate on radio phone-ins and in the local newspaper had been ` which of the two matches would you prefer to win ? ' Losing both was never an option .Pardew , of course , has done tremendously well to transform his own and Newcastle 's fortunes , but the mood can change quickly on Tyneside .After this humiliation , Pardew will be only too aware of Sunday 's significance .\\

\midrule
\textbf{Fine-tune LLaMa on 5W1H extraction} \\
\midrule
\textcolor{red}{"what"}: ["Debate has raged in Newcastle as to which their fans would rather see - a place in the Capital One Cup semi-final or a first win over Sunderland in six attempts . The majority sided with the former and a shot at a first domestic trophy since 1955 .That dream , however , is dead"],\\ \textcolor{yellow}{"when"}: ["on Wednesday, 1955"],\\ \textcolor{blue}{"where"}: [" Newcastle, Tottenham, London"],\\ \textcolor{green}{"why"}: [" The majority sided with the former and a shot at a first domestic trophy since 1955 .That dream , however , is dead"],\\ \textcolor{orange}{"who"}: ["Alan Pardew, Massadio Haidara, keeper Jak Alnwick, Newcastle, Moussa Sissoko, Yoan Gouffran, Remy Cabella, Emmanuel Riviere, Christian Eriksen, Nabil Bentaleb, Gus Poyet, Jordi Gomez"],\\ \textcolor{purple}{"how"}: ["Well , it just got bigger .Massadio Haidara consoles keeper Jak Alnwick after a night to forget for Newcastle at Tottenham .Newcastle 's defence reels as Spurs celebrate the final goal of their 4-0 defeat of the Toon on Wednesday"]\\

\midrule
\textbf{Few-shot on ChatGPT} \\
\midrule
\textcolor{yellow}{When}: Sunday, May 12\\ \textcolor{blue}{Where}: Tyne-Wear derby in Newcastle\\ \textcolor{green}{Why}: To determine whether Newcastle would rather advance to the Capital One Cup semi-final or win over Sunderland\\ \textcolor{orange}{Who}: Alan Pardew (Newcastle manager), Moussa Sissoko (Newcastle player), Yoan Gouffran, Remy Cabella, Emmanuel Riviere (Newcastle players), Jak Alnwick (Newcastle goalkeeper), Christian Eriksen and Nabil Bentaleb (Spurs players), Roberto Soldado (Spurs player), Gus Poyet (Sunderland manager)\\ \textcolor{purple}{How}: Newcastle lost 4-0 to Spurs in the Capital One Cup, causing them to focus on their upcoming match against Sunderland\\

\midrule
\textbf{Zero-shot on GPT-4} \\
\midrule
\textcolor{red}{What}: The upcoming Tyne-Wear derby between Newcastle and Sunderland. \\ \textcolor{yellow}{When}: Sunday (No specific date given)\\ \textcolor{blue}{Where}: Newcastle\\ "Why": The significance of the Tyne-Wear derby has increased after Newcastle's loss in the Capital One Cup semi-final. 
\\textcolor{orange}{Who}: Newcastle fans, Manager Alan Pardew and his players, Sunderland team, Spurs side\\ \textcolor{purple}{How}: Newcastle lost against Spurs which resulted in their ousting from the Capital One Cup. This has increased the pressure on them for their upcoming match against Sunderland.\\
\midrule
\textbf{Few-shot on GPT-4} \\
\midrule
\textcolor{red}{"what"}: ["Sunday 's Tyne-Wear derby takes on a new dimension now they 're out of the cup", "It was , in the end , a rather timid surrender", "The thought of going into that game without Sissoko is a frightening one"], \\ \textcolor{yellow}{"when"}: ["on Wednesday", "in five days", "this weekend", "Sunday"],\\ \textcolor{blue}{"where"}: ["Newcastle", "at Tottenham", "in north London"],\\ "why": ["Alan Pardew and his players had talked of a ` massive ' week in their season . Well , it just got bigger"],\\ 
\textcolor{orange}{"who"}: ["Massadio Haidara ", "Jak Alnwick","Alan Pardew's side ","Moussa Sissoko "],\\ \textcolor{purple}{"how"}:["The manager had , in fairness , named just about the strongest XI available to him but it looked awfully weak against a Spurs side who were superior in every department"]\\

\bottomrule
\bottomrule

\end{tabular}
}

% \caption{Examples of 5W1H element extraction in different situations. The source document is annotated by annotators.}
\label{table4}%

\end{table*}
\paragraph{\bf Adaptation Between Different Datasets}Table \ref{tab:addlabel3} shows the extraction results of the model on the other two datasets. The first line represents the result after fine-tuning with its own dataset, and the second line represents the model after fine-tuning on CNN/DailyMail in other datasets.
We find that statements about \textit{Who}, \textit{Where}, and \textit{When}, which are relatively concise in their answers, don't show significant variations in their effectiveness across different datasets. On the other hand, the extraction of \textit{What}, \textit{Why}, and \textit{How}, which requires a more comprehensive understanding of the article, proves to be less effective in comparison. We think that increasing the training dataset such as integrating the training of these four labeled datasets may have an improvement in the model's extraction performance.
\paragraph{\bf Example Analysis}
Table \ref{table4} shows the comparison between the fine-tuned model, ChatGPT and GPT-4. 
For short entities, the fine-tuned LLaMa model can extract some entities from the original text but may still miss some cases. For long entities, due to the impact of model performance and the size of the fine-tuning dataset, there are still some repetitions in the answers to \textit{Why} and \textit{What}. ChatGPT performs not well after being provided with examples and does not learn the output format. In contrast, The output of GPT-4 is similar to manually annotated.

\section{CONCLUSION}
In this work, we annotate 5W1H elements from the text to address the extraction challenges and fine-tune the datasets using LLMs. The extraction performance is comparable to GPT-4's few-shot capability and outperforms ChatGPT. We also find that our annotated datasets can express their ability to extract 5W1H elements across different news domains by increasing model parameters and training data. Therefore, creating a high-quality dataset is crucial in addressing challenges faced by LLMs in news extraction.

\section{Acknowledgements}
We extend our gratitude to all the reviewers for their professional advice. This research is supported by the National Natural Science Foundation of China (No.62106105), the CCF-Baidu Open Fund (No.CCF-Baidu202307), the CCF-Zhipu AI Large Model Fund (No.CCF-Zhipu202315), the Scientific Research Starting Foundation of Nanjing University of Aeronautics and Astronautics (No.YQR21022), and the High Performance Computing Platform of Nanjing University of Aeronautics and Astronautics.

\bibliographystyle{IEEEtran}
\bibliography{IEEEabrv,reference}
\end{document}